\documentclass[sigconf]{acmart}

\usepackage{todonotes}
\usepackage{booktabs}
\usepackage{graphics}
\usepackage{xspace}
\usepackage{amsmath}
\usepackage{caption}
\usepackage{subcaption}
\DeclareMathOperator*{\argmax}{arg\,max}

\AtBeginDocument{%
  }

\renewcommand\footnotetextcopyrightpermission[1]{}
\newcommand\blfootnote[1]{%
\begingroup
\renewcommand\thefootnote{}\footnote{#1}%
\addtocounter{footnote}{-1}%
\endgroup
}

\begin{document}

\title{Reinforcement Learning-based Placement of Charging Stations in Urban Road Networks}

\author{Leonie von Wahl}
\affiliation{%
  \institution{Volkswagen Group}
  \streetaddress{}
  \city{Hannover}
  \country{Germany}}
\email{leonie.von.wahl@volkswagen.de}
\orcid{0000-0003-0013-831X}

\author{Ashutosh Sao}
\affiliation{%
  \institution{L3S Research Center,  University of Hannover}
  \streetaddress{}
  \city{Hannover}
  \country{Germany}}
\email{sao@L3S.de}
\orcid{0000-0002-7049-3387}

\author{Nicolas Tempelmeier}
\affiliation{%
  \institution{Volkswagen Group}
  \streetaddress{}
  \city{Hannover}
  \country{Germany}}
\email{nicolas.tempelmeier@volkswagen.de}
\orcid{0000-0003-0911-6264}

\author{Elena Demidova}
\affiliation{
  \institution{Data Science \& Intelligent Systems Group (DSIS)}
  \streetaddress{}
  \city{University of Bonn, Bonn}
  \country{Germany}}
\email{elena.demidova@cs.uni-bonn.de}
\orcid{0000-0002-5134-9072}

\renewcommand{\shortauthors}{von Wahl et al.}

\begin{abstract}
 The transition from conventional mobility to electromobility largely depends on charging infrastructure availability 
 and optimal placement.
This paper examines the optimal placement of charging stations in urban areas. 
  We maximise the charging infrastructure supply over the area and minimise waiting, travel, and charging times while setting budget constraints.
  Moreover, we include the possibility of charging vehicles at home to obtain a more refined estimation of the actual charging demand throughout the urban area.  
  We formulate the Placement of Charging Stations problem as a non-linear integer optimisation problem that seeks the optimal positions for charging stations and the optimal number of charging piles of different charging types. 
  We design a novel Deep Reinforcement Learning approach to solve the charging station placement problem (PCRL). Extensive experiments on real-world datasets show how the PCRL reduces the waiting and travel time while increasing the benefit of the charging plan compared to five baselines. Compared to the existing infrastructure, we can reduce the waiting time by up to 97\% and increase the benefit up to 497\%.
\end{abstract}




\maketitle
\pagestyle{plain}
\blfootnote{\textcopyright Leonie von Wahl, Nicolas Tempelmeier, Ashutosh Sao, Elena Demidova 2022. This is the author's version of the work. It is posted here for your
personal use. Not for redistribution. The definitive version was published in the proceedings of the 28th ACM SIGKDD Conference on Knowledge Discovery and Data Mining (KDD '22) \url{https://doi.org/10.1145/3534678.3539154}.\\}

\section{Introduction}
\label{sec:intro}

Electromobility has developed as indispensable transportation means for modern transportation systems.
To reduce greenhouse gas emissions and to counteract the climate crisis, many governments provide 
incentives to stimulate broader electromobility adoption \cite{XU2020102534}.
The provision of accessible charging infrastructure is the critical requirement to reduce range anxiety (i.e. the driver's concerns that a vehicle energy storage is insufficient to reach the trip destination) and enable the widespread adoption of electric vehicles \cite{GREENE2020102182}.
Whereas the number of charging stations (CS) has grown recently, the currently available number of CS is insufficient to satisfy future charging needs.
However, there is a lack of methods to automatically determine appropriate new CS locations.

The problem of optimally placing CS within a road network, i.e., determining the required capacity and position, is particularly challenging due to the following factors:
First, numerous factors influence the optimal placement, including road network topology, existing charging infrastructure, traffic patterns, and charging duration.
Second, the charging demand typically covers the whole road network such that regional optimisation methods, considering only isolated junctions or parking lots, fail to determine a suitable placement.
Existing approaches for CS placement rely on simple greedy algorithms that fail to address the complex spatio-temporal interdependencies resulting from the road network topology and the charging demand \cite{10.1145/3347146.3359382}. Consequently, they tend to cluster the CS, leading to high travel times and ultimately failing to address real-world needs for electric vehicle drivers. 

In this paper, we address the problem of CS placement by introducing a novel formulation of the CS placement problem. 
Our utility model considers the basic supply with charging infrastructure and the driver's discomfort.
We explicitly consider the possibility of home-charging electric vehicles \cite{FUNKE2019224}, an essential aspect currently neglected in the prior work.
The home-charging eases the problem of an insufficient public charging infrastructure \cite{ZHAO2020120039} as electric vehicles are likely often charged in residential areas.

Moreover, we present a novel formulation of the \textbf{p}lacement of \textbf{c}harging stations task as a \textbf{r}einforcement \textbf{l}earning problem (PCRL) based on a deep Q network algorithm.
We evaluate the PCRL on four real-world datasets and demonstrate that our proposed method systematically determines effective CS placement. The evaluation against the existing CS placement models shows that our model creates a superior charging infrastructure regarding the general supply as well as waiting and travel times. 
Furthermore, we provide evidence that automated placement of CS is feasible and that considering home-charging facilities helps to obtain functional real-world solutions.
Finally, we discuss the prerequisites for deploying CS allocation models regarding data needs and required interfaces.

In summary, our contributions are as follows:
\begin{itemize}
    \item We propose a novel formal model for optimal charging infrastructure placement in a road network. 
    Our utility model improves over established problem formalisations by considering the home charging of electric vehicles.
    \item We formulate CS placement as a reinforcement learning problem. 
    We propose novel action and observation spaces to determine efficient CS placements automatically.
    \item We conduct extensive experiments on real-world datasets, demonstrating the performance of our approach in comparison against five baselines. 
    We provide evidence indicating the potential benefit of the deployment of automated charging station placement models.
\end{itemize}

Compared to the existing infrastructure, we can reduce the waiting time by up to 97\% and increase the benefit up to 497\%.
For reproducibility, our code is publicly available\footnote{Our code is publicly available at: \url{https://github.com/ashusao/PCRL.git}}.

\section{Problem Statement}
\label{sec:problem}

In this section, we formally introduce the charging station placement problem.
First, we model a road network as a directed graph.
\begin{definition}[Road Network]
Let $G = (V, E)$ be a directed weighted graph with $V$ the set of vertices and $E$ the set of edges. 
The vertices are the road network junctions, while the edges represent the roads direction-wise. 
\end{definition}

\begin{figure}[t]
	\centering
  \includegraphics[width=8cm]{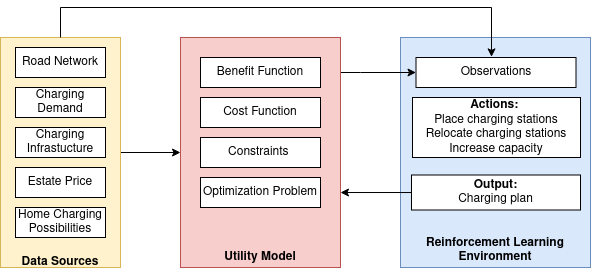}
	\caption{Overview of data sources, optimisation problem and approach proposed in this paper.}
	\label{Fig:Scheme}
\end{figure}

Next, we introduce charging stations located within the road network $G$.
\begin{definition}[Charging Station]
\label{def:station}
A \textit{charging station} $s$ within the road network $G$ is defined as a tuple $s=(v, t)$, where $v \in V$ is the location node, $t=(t_1, ..., t_m)$ is a vector of length $m \in \mathbb{N}$ with $t_{i} \in \mathbb{N}$ being the number of chargers of type $i$ at $s$.
We denote the set of all possible CS as $S$.
\end{definition}

The overall placement of CS within $G$ forms a charging plan.
\begin{definition}[Charging Plan]
\label{def:plan}
A \textit{charging plan} $p$ on $G$ is an assignment of vertices to charging stations. Formally, we define a charging plan $p \subset S$ as a CS set that includes at most one CS per vertex.
We denote the set of all possible charging plans (CP) as $P$.
\end{definition}

We quantify the advantages and drawbacks of a charging plan through a benefit and a cost function.
The $\emph{benefit}: P \to \mathbb{R}$ of a CP $p$ quantifies the positive effect of $p$ for satisfying the charging demand.
The $\emph{cost}: P \to \mathbb{R}$ of $p$ quantifies the effort required to realise $p$. 
The \emph{score} expresses the overall quality of a CP:
\begin{equation}
\label{eq:score}
    score(p) = \lambda \cdot benefit(p) - (1-\lambda) \cdot cost(p).
\end{equation}
Here, $\lambda \in [0,1]$ is a weighting parameter to trade off $benefit(p)$ and $cost(p)$.
Finally, we formulate the CS placement problem:
\begin{definition}[Charging Station Placement Problem]
Given a road network $G$, a \emph{benefit} function, and a \emph{cost} function, find the optimal charging plan 
\begin{equation}
p^* = \argmax_{p \in P}\{ \lambda \cdot benefit(p) - (1-\lambda) \cdot cost(p)\}
\label{eq:problem}
\end{equation}  
that maximises the overall utility by trading off $benefit(p)$ and $cost(p)$.
\end{definition}

\section{Utility Model}
This section quantifies the utility of a particular CP $p$ for a road network $G$.
To this end, we define a $benefit$ and a $cost$ function for $p$.
In this section, we extend the current state-of-the-art formalisation for CS placement \cite{10.1145/3347146.3359382}.

\subsection{Benefit Function}
We formulate the benefit function based on the following intuition:
\begin{itemize}
    \item CS with a higher capacity should result in a higher benefit since they can provide service for a higher number of vehicles.
    \item Existing nearby CS should decrease the benefit of a new CS since existing demands are already partially satisfied.
\end{itemize}

\emph{Charging station capacity.} Let $s=(v,t)$ be a CS with $m$ charger types summarised by the numbers of chargers $t=(t_1, ..., t_m) \in \mathbb{N}^m$.
Let $c_i$ denote the available charging power of the charger $t_i$. 
We define the capacity $C(s)$ of a CS $s$ as the sum of the charging power provided by the individual chargers at $s$ with
$    C(s) = \sum_{i=1}^m t_{i} c_i $.

\emph{Influential radius \& coverage.} The influential radius expresses the distance in which the CS attracts electric vehicles. 
Intuitively, a higher capacity should result in a higher influential radius. We denote the maximal influential radius as $r_{max}$.
Formally, given a CS $s$ with scaled down and dimensionless capacity $\widetilde{C}(s)$, we define the \emph{influential radius} $r(s)$ as
\[r(s) = r_{max}  \frac{1}{ 1 + \exp(-\widetilde{C}(s))}.\]
The \emph{coverage} $\textit{cov}(v)$ of a vertex $v$ is defined as the number of CS in whose influential radius $v$ is:
$\textit{cov}(v) = |\{s \in P | d(v,s) \le r(s)\}|$. Here, $d(v,s)$ denotes the haversine distance between the CS $s$ and the junction $v$. Hence, the coverage of a node describes how good it is supplied with charging stations. 
The coverage $\textit{cov(v)}$ is less critical in areas where there is much opportunity for home charging, e.g. by CS in private garages. Therefore, we establish $home(v) \in [0, 1]$ as the share of detached houses among all buildings located around $v$.
Then, we formulate the overall benefit function of a CP $p$:

\begin{equation}
\label{eq:benefit}
    benefit(p) = \frac{1}{|V|}\sum_{v \in V} \left(\sum_{i=1}^{cov(v)} \frac{1}{i} \right) (1 - \omega \; home(v)).
\end{equation}
Here, $\omega$ is a weighting parameter. Intuitively, the benefit expresses the supply of the road network with electric charging infrastructure. A higher coverage leads to a higher benefit. The formulation above ensures that it is more beneficial to cover different vertices with the CS instead of clustering all CS in a dense road network area.

\subsection{Cost Function}
We formulate the cost of a CP $p$ based on the expected required time to charge a vehicle.
To this end, we consider \emph{travel time}, \emph{charging time}, and \emph{waiting time} (if all chargers are occupied).

To estimate the travel time, we consider the distances between
the junctions of the road network and the respective nearest CS
and the charging demand at the individual junctions. The demand is included since the travel time contains all travel times according to the frequency of their occurrence. 
Let $dem(v) \in [0, 1]$ denote the charging demand of a junction, i.e., the normalised number of vehicles that are typically parked at $v$ and require charging.

Since the charging demand is partly satisfied by the home charging infrastructure, we introduce a so-called \emph{weakened demand} as
\begin{equation*}
    dem_{weak}(v) = dem(v)(1 - \omega \, home(v)).
\end{equation*}

We define the indicator function $i:S \times V \mapsto \{0,1\}$ to be $1$ for pairs of CS $s \in S$ and junctions $v \in V$ if $s$ is the CS assigned to the junction $v$ according to the CS assignment scheme in \cite{10.1145/3347146.3359382}, and $0$ otherwise.
We then define the over travel time induced by a CP $p$ on a road network with junctions $V$ as:
\[
    \textit{travel}(p) = \sum_{v \in V} \sum_{s \in p} i(s,v) \frac{dist(s,v)} {\mathcal{V}} dem_{weak}(v)
\]
where $dist(s,v)$ is the haversine distance and $\mathcal{V} = const.$ is the average velocity in town. 

Next, we model the \emph{charging time}. We estimate the number of vehicles approaching a CS $s$ as:
\[
    D(s) = \sum_{v \in V} \frac{i(s,v)}{dist(s,v)} dem_{weak}(v).
\]
The \textit{service rate} $\mu(s)$ of a station is 
\[
    \mu(s) = C(s) / E.
\]
Here $E$ is the energy required to charge a single-engine.
We then define the overall charging time for a CP $p$ as:
\[
    charging(p) = \sum_{s \in p} \frac{D(s)}{\mu(s)}.
\]

We model the \textit{expected waiting time} $W(p)$ using the
Pollaczek-Khintchine formula \cite{10.1145/3347146.3359382}:
\[
    waiting(p) = \sum_{s \in p} W(s) D(s)
\]
with
\[
    W(s) = \frac{\rho(s)}{2 \mu(s) (1 - \rho(s))}, \text{ for } \rho(s) < 1,
    \quad \text{ and } \quad
    \rho(s) = \frac{D(s)}{\mu(s)}.
\]

Finally, we combine travel time, charging time and waiting time of a CP $p$ into a single formula
\[
    cost(p) = \alpha travel(p) + (1 - \alpha) (charging(p) + waiting(p)),
\]
where $\alpha \in [0,1]$ is a weighting parameter.

\subsection{Constraints \& Optimization}
To facilitate the identification of practical charging plans, we extend the problem formulation of Equation \ref{eq:problem} by several constraints.
We formulate the constrained non-linear integer optimisation problem to find the optimal CP $p^*$ in Equations \ref{eq:Aim1}-\ref{eq:Aim5} as follows:

First, we limit the financial costs of $p^*$ by a fixed budget $B$ (\ref{eq:Aim2}) explained in detail in the following.
Next, we limit the number of installed chargers at a single CS by the constant $K$ (\ref{eq:Aim4}).
Finally, we ensure that the formulation of the waiting time is well-defined and positive (\ref{eq:Aim5}).
\begin{align}
\label{eq:Aim1}
&p^* = \argmax_{p \in P}\{ \lambda \cdot benefit(p) - (1-\lambda) \cdot cost(p) \} \\\
\label{eq:Aim2}
\text{s.t. } \; & \sum_{s \in p*} \textit{fee(s)} \leq B \\
\label{eq:Aim4}
& \sum_{i=1}^m s.t_i \le K \quad \forall s \in p*\\
\label{eq:Aim5}
& \rho(s) < 1 \quad \forall s \in p*.
\end{align}

\emph{Installation costs:} We consider the financial cost of installing new CS for our budget constraint in Equation \ref{eq:Aim2}.
Given a CS $s$, the \textit{instalment fee} denoted by $\text{fee}(s)$ is the estate cost at the node $s.v$ $\textit{estate-cost}(s.v)$ and the number of installed chargers of type $i$ and their corresponding installation costs $\textit{charger-cost}(i)$.
\begin{equation*}
    \textit{fee}(s) = \textit{estate-cost}(s.v) + \sum_{i=1}^m s.{t_i} \cdot \textit{charger-cost}(i).
\end{equation*}

\section{Reinforcement Learning}
\label{sec:rl}

This section describes the optimisation problem stated above as a reinforcement learning (RL) problem. 
We provide a brief description of our implementation.
\subsection{RL Problem Formulation}
We model the problem of CS placement as a \emph{single agent} reinforcement learning problem (PCRL).
In the following, we detail the PCRL formulation in terms of observations, actions, rewards, and episodes, where $i \in \mathbb{N}$ denotes the number of the episode.

\textbf{Observations.}
An observation 
$o^i \in P \times \mathbb{R}^{|V|^2} \times \mathbb{R}^{|V|} \times \mathbb{R}^{|V|} \times \mathbb{R}^{|V|}$ captures the intermediate placement of the CS and consists of the current CP, the latitude and longitude coordinate, the resulting charging demand $dem(v)$, the private charging infrastructure $priv(v)$ and the estate price \textit{estate-cost}$(v)$ for each node $v \in V$.

\textbf{Actions.}
We use a set of discrete actions, i.e., an action $a^i \in \{\textit{Create by} \textit{ benefit}, \textit{Create} \textit{ by demand}, \textit{Increase} \textit{ by benefit}, \textit{Increase} \\ \textit{ by demand}, \textit{Relocate}\}$. 
The agent chooses the action from the set of discrete actions and can either \emph{create} a new CS, \emph{increase} the number of chargers at an existing CS or \emph{relocate} a charger from one CS to another. We relocate CS instead of deleting them because it stimulates the agent to build more stations. 

For \emph{creating} a new CS, we employ one of two different greedy strategies to determine the position of the CS $s$ within the road network. If the agent chooses the action \textit{Create by benefit}, it will choose the node with the lowest coverage $cov(v)$. 
The idea is to increase the probability that at least one CS covers each node.
When the agent chooses \textit{Create by demand}, we choose the node with the highest weakened demand $dem_{weak}(v)$. To select the chargers at the station, we create a lookup table before the training starts: We calculate the capacity of each feasible charging configuration out of the $m$ charger types. Then, we sort the configurations and write the cheapest charger configuration for each possible capacity demand into the lookup table. Now, for the action, we estimate the required capacity of the candidate station and then assign the cheapest charger configuration to it.

Similar, for \emph{increasing} the number of chargers, we employ the same two greedy strategies as above to select a CS $s \in P$. At that position, we add one charger. We can only add a charger to a station with less than $K$ chargers. 

For \emph{relocating} chargers, we determine the CS $s_{old}$ achieving the lowest benefit.
Then, we relocate one of its chargers to the CS with the highest waiting and charging time in the current CP $p^i$. The idea is to support stations with an unusually high waiting time.
If $s_{old}$ is left without any chargers, we remove $s_{old}$ from $p^i$. We assume that the current supply is lower than the entire demand, such that a relocation instead of a deletion is always desirable. Currently, we neglect the cost of relocation of existing charging infrastructure.

\textbf{Rewards.}
We calculate the reward based on the proposed utility function.
To this end, we compare the current CP $p^i$ with the CP resulting from action $a^i$ using the $score(p)$ function. The initial score is zero if we start with an empty CP. If we extend the existing charging infrastructure, we initiate the training with the score of the original CP.
We combine the score difference to obtain the reward function:
\[
r^i = score(p^{i+1}) - score(p^{i}).
\]

\textbf{Episodes.}
An episode starts with an initial CP $p^0$ that can either
be empty or represent the existing charging infrastructure.
The episode ends if one of the following conditions holds:
\emph{Budget exceeded}: The current action $a^i$ requires more budget than what is left from the starting budget $B$.
\emph{Maximum number of chargers reached}: The current CP $p^i$ assigns $K$ charging stations to all junctions $v \in V$, so there is no more place for any new charger.
\emph{Maximum iteration reached:} The current iteration number $i$ reaches the maximum number of iterations  $i_{max}$, see Sec. \ref{sec:parameters}.

\subsection{RL Implementation}
To solve the reinforcement learning problem, we apply a deep Q net-
work reinforcement learning \cite{mnih2013playing}.
Deep Q learning combines a convolutional neural network trained via stochastic gradient descent with an experience replay mechanism to address data correlation and non-stationary distributions.
We employ the implementation provided by the Stable Baselines3 framework \cite{stablebaselines3}. 

\section{Evaluation Setup}
\label{sec:setup}

In this section, we describe the model parameter, baselines, datasets and evaluation metrics. 

\subsection{Model parameter}
\label{sec:parameters}
In the utility model, we apply the following values for the parameters: $\alpha = 0.4$, $m = 3$, $K = 8$, $B =$ €5 Mio., $E = 85$kWh, $r_{max} = 1$km, $w$=0.1. We set $\lambda = 0.5$ to prioritise the cost function and the benefit function equally. The charging power $c_1, c_2, c_3$ for the three charger types is 7 kW, 22 kW, 50kW. The respective \textit{estate-cost} is €300, €750, €28000. We base our prices on the usual manufacturer prices.

When implementing the PCRL as deep Q learning model, we set \textit{policy} = \textit{MlpPolicy}, \textit{batch-size} = 128, \textit{buffer-size} = 10000 and \textit{learning-rate} = 0.001. Moreover, we set $i_{max} = \frac{\# nodes}{2}$ depending on the underlying road network. We train for up to 200,000 episodes.
To ensure the reproducibility of our results, we set a random seed.

\subsection{Baselines}
\label{sec:baselines}

\newcommand{\existingChargers}{\textsc{Existing Charging}\xspace}
\newcommand{\boundOpt}{\textsc{Bounding}\&\textsc{Optimising}\xspace}
\newcommand{\boundOptPlus}{\boundOpt\textsc{+}\xspace}
\newcommand{\bestBenefit}{\textsc{BestBenefit}\xspace}
\newcommand{\highestDemand}{\textsc{Highest Demand}\xspace}
\newcommand{\chargerGreedy}{\textsc{Charger-based Greedy}\xspace}

To evaluate our solution to the problem formulated in Eq. \ref{eq:Aim1} we compare it to several na\"ive and state-of-the-art baselines.

\existingChargers. 
We examine the status quo of charging infrastructure, i.e., the public CS already existing in the road network. 
Comparing our approach to this baseline provides an intuition about the added value of the CP computed by our approach.

\boundOpt.
This  baseline is the \textit{Bounding \& Optimising Based Greedy} proposed in \cite{10.1145/3347146.3359382}. 
We choose a new CS position by comparing the potential benefit of all potential positions. 
An unbounded knapsack algorithm delivers an initial charger configuration. We add more chargers to the CS according to the constraints. Then, we include the new CS in the CP.

\boundOptPlus. 
We implement a \textit{Bounding \& Optimising Based Greedy Plus} baseline as an improved version of the \textit{Bounding \& Optimising Based Greedy} algorithm, making sure that the resulting CP is a feasible solution to the PCS.
We replace the unbounded knapsack algorithm with the same algorithm as in the PCRL action space to get an initial charging configuration.

\bestBenefit.
This greedy baseline implements a heuristic that always chooses the junction with the highest possible $benefit$ as the CS. 
We use the same initial charging configuration as in the \boundOptPlus. 

\highestDemand.
Similar to \bestBenefit, this baseline places the next CS at the junctions with the highest demand.

\chargerGreedy.
Du et al. \cite{10.1145/3219819.3220032} proposed the \textit{Fast Charger-based greedy algorithm}. 
This baseline uses its own utility model that considers points of interest, the local charging demand and other information defined in detail in \cite{10.1145/3219819.3220032}.
In each iteration, the baseline places new CS or increases the capacity of existing CS greedily.
In particular, \chargerGreedy\text{ } always chooses the action that increases its utility score the most.

\subsection{Datasets}
\label{sec:Datasets}

\begin{figure*}
     \centering
     \begin{subfigure}[]{0.32\textwidth}
          \centering
     \includegraphics[width=5.0cm]{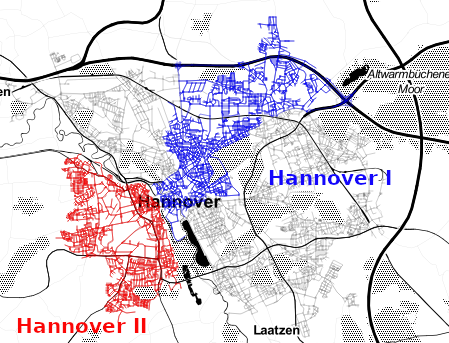}
     \caption{Road networks Hannover I and II.}
     \label{Fig:hannover}
     \end{subfigure}
         \begin{subfigure}[]{0.32\textwidth}
          \centering
     \includegraphics[width=4.0cm]{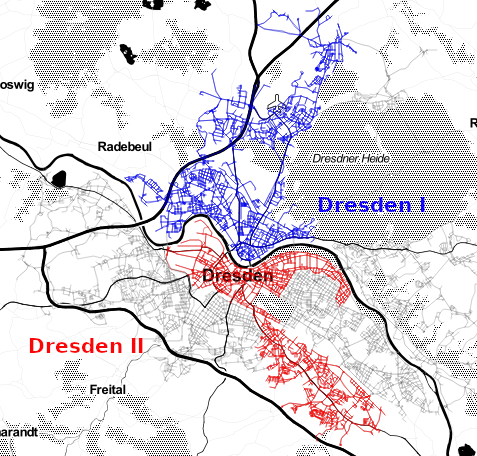}
     \caption{Road networks Dresden I and II.}
     \label{Fig:dresden}
     \end{subfigure}
       \caption{Road networks in Hannover and Dresden datasets.}
     \end{figure*}

\paragraph{Road Networks}
We evaluate our approach on four equally sized road networks from the two German cities, Hannover and Dresden, named Hannover I, Hannover II, Dresden I, and Dresden II. We choose the datasets to vary in charging demand, estate price and density of detached houses.
The road networks are depicted in Figure \ref{Fig:hannover} and \ref{Fig:dresden}. Tabular \ref{Tab:Dataset} provides selected dataset statistics.

\emph{Hannover I} consists of the three districts Mitte, Vahrenfeld-List and Bothfeld-Vahrenheide. 
For \emph{Hannover II}, we use the districts Linden-Limmer, Ricklingen and Ahlem-Badenstedt-Davenstedt. 
\emph{Dresden I} consists of the districts Weixdorf, Klotzsche, Pieschen and Neustadt. 
\emph{Dresden II} is composed of Altstadt, Prohlis and Blasewitz.

\paragraph{Charging Demand}
To compute the charging demand, we use a proprietary traffic dataset of Hannover and Dresden from 15 Aug 2019 to 01 Dec 2019. 
We divide the datasets into $32$ x $32$ grid cells and count the number of trips ending in individual grid cells for the period. 
We expect that a high number of trips ending in a cell, i.e., a high number of parked vehicles, corresponds to a high demand.
Our approach can easily be reproduced with other traffic datasets.

\paragraph{Charging infrastructure}
We receive the locations of the charging stations already built in Hannover from the Open Charge Map portal \cite{OCM}. This portal provides information about CS containing location and the total number of ports available with the port type. 

Since it is impossible to get information on the exact number of private home garages, we use the density of detached houses to estimate the impact of private charging infrastructure.
We extract from OpenStreetMap \cite{OSM} how much of the city area is used as residential land. From the data of the city municipality, we receive the share of detached houses among all buildings for each district, respectively. Hence, we can attribute a value between 0 and 1 to each road network node. This value indicates the density of potential home charging infrastructure around this node.

\paragraph{Estate price}
We extract the estate price data from the city municipality's rent index 2021. Thus, we can attribute each district's estate price per square meter.

\begin{table}[t]
\centering
\caption{Statistics of the datasets.}
\label{Tab:Dataset}
\resizebox{\columnwidth}{!}{%
 \begin{tabular}{lcccc} 
\toprule
& Hannover I & Hannover II  & Dresden I & Dresden II \\
\midrule
\# nodes & 1947 & 1707 & 1919 & 1897 \\
\# exist. CS & 55 & 50 & 29 & 27 \\
\# of edges & 4766 & 4121 & 4849 & 4572 \\
$\overline{node degree}$ & 2.85 & 2.68 & 2.85 & 2.83 \\
\bottomrule
\end{tabular}
}
\end{table}

\subsection{Evaluation Metrics}
\label{sec:setup:metrics}
To understand the goodness of the charging plan, we propose the following quality indicators:\\
\textbf{Score ($score$).} 
The overall score of the CP according to the utility model as defined by Eq. \ref{eq:score}.
The score quantifies the overall performance of the respective model (higher is better).
\\
\textbf{Benefit ($benefit$).}
The benefit of the CP, i.e., the score without the cost function, as defined by the utility model in Eq. \ref{eq:benefit}.
The benefit quantifies the potential usefulness of the respective model (higher is better).
\\
\textbf{Waiting Time ($wait$).}
The sum of the waiting times occurring at all CS in the road network (lower is better). 
\\
\textbf{Travel Time ($travel$).}
The sum of travelling times within the road network (lower is better).
\\
\textbf{Charging Time ($charging$).}
The sum of charging times within the road network (lower is better).
\\
\textbf{Maximum Travel Time ($travel_{max}$).}
The maximum time it takes to travel to reach a CS from any junction in the road network any CS (lower is better).
\\
\textbf{Maximum Waiting Time ($wait_{max}$).}
The maximum time spent waiting at any CS in the road network (lower is better).

To enable the comparison to the existing real-world charging infrastructure, we report the benefit, waiting time, and charging time relative to the performance of the \existingChargers.
We consider the performance of \existingChargers as 100\% and provide the performance of all other models as percentages relative to it.
We report the maximum travel time and the maximum waiting as absolute numbers to provide intuitive measurements.

\section{Evaluation}
\label{sec:eval}

The evaluation aims to assess the performance of our developed PCRL approach on real-world datasets.

First, we analyse the deployment of new charging stations in a real-world scenario in Section \ref{sec:eval:real}.
To this end, we consider the currently existing charging stations as an initial charging plan and incrementally add new charging stations.
In the second experiment, we investigate the principal model behaviour when the model is not restricted to improving the existing charging infrastructure in Section \ref{sec:eval:new}. 
Finally, we discuss the prerequisites and opportunities for the deployment of our approach in Section \ref{sec:eval:deployment}.

\begin{table*}[t]
\centering
\caption{Results on the Hannover datasets. Evaluation metrics where higher values are better are marked with $\uparrow$. Metrics where lower values are better are labelled with $\downarrow$. Best scores are marked bold. ``-'' indicates that no valid solution was found.}
\label{Tab:Hannover}
\resizebox{310pt}{!}{%
\begin{tabular}{l cc cccccc} 
 \toprule
 \multicolumn{1}{c}{Algorithm} & \multicolumn{2}{c}{Score $\uparrow$} & \multicolumn{5}{c}{Cost $\downarrow$} \\
 \cmidrule(lr){2-3}\cmidrule(lr){4-8}
 & $score$ & $benefit$ & $wait$ & $travel$ & $charging$ &  $travel_{max}$ [min] & $wait_{max}$ [min]\\
\midrule
\multicolumn{8}{c}{Hannover I}  \\
\cmidrule(lr){1-8}
\existingChargers & 100\% & 100\% & 100\% & 100\% & 100\% & 7.27 & 32.38\\
\boundOpt & - & - & - & - & - & - & - \\
\boundOptPlus & 284\% & 171\% & 47\% & 91\% & 85\%  & 7.27 &  22.46 \\
\bestBenefit & 268\% & 164\% & 49\% & 92\% & 87\% & 7.27  & 22.46\\
\highestDemand & 233\% & 143\% & 39\% & 81\% & \textbf{80\%}  & 7.27 & 21.83 \\
\chargerGreedy &  17\% & 136\% & 80\% & 98\% & 123\% & 7.27 & 22.46 \\
PCRL & \textbf{376\%} & \textbf{202\%} & \textbf{18\%} & \textbf{57\%} & \textbf{80\%} & \textbf{5.93} & \textbf{8.11}\\
\midrule
\multicolumn{8}{c}{Hannover II}  \\
\cmidrule(lr){1-8}
\existingChargers & 100\% & 100\% & 100\% & 100\% & 100\% & 15.94 & 13.96 \\
\boundOpt & - & - & - & - & - & - & - \\
\boundOptPlus & 353\% & 204\% & 61\% & 87\% & 78\% & 15.94 & 13.96 \\
\bestBenefit & 329\% & 195\% & 63\% & 86\% & 88\% & 15.94 & 13.96 \\
\highestDemand & 262\% & 155\% & 36\% & 70\% & 82\% & \textbf{15.62} & 7.42 \\
\chargerGreedy & 145\% & 121\% & 100\% & 100\% & 101\% & 15.94 & 13.96  \\
PCRL & \textbf{428\%} & \textbf{222\%} & \textbf{10\%} & \textbf{46\%} & \textbf{69\%} & 15.63 & \textbf{1.37} \\
\bottomrule
\end{tabular}
}
\end{table*}

\begin{table*}[t]
\centering
\caption{Results on the Dresden datasets. Evaluation metrics where higher values are better are marked with $\uparrow$. Metrics where lower values are better are labelled with $\downarrow$.  Best scores are marked bold. ``-'' indicates that no valid solution was found.}
\label{Tab:Dresden}
\resizebox{310pt}{!}{%
\begin{tabular}{l cc cccccc} 
 \toprule
 \multicolumn{1}{c}{Algorithm} & \multicolumn{2}{c}{Score $\uparrow$} & \multicolumn{5}{c}{Cost $\downarrow$} \\
 \cmidrule(lr){2-3}\cmidrule(lr){4-8}
 & $score$ & $benefit$ & $wait$ & $travel$ & $charging$ &  $travel_{max}$ [min] & $wait_{max}$ [min]\\
\midrule
\multicolumn{8}{c}{Dresden I}  \\
\cmidrule(lr){1-8}
\existingChargers & 100\% & 100\% & 100\% & 100\% & 100\% & 27.30 & 42.52 \\
\boundOpt & - & - & - & - & - & - & - \\
\boundOptPlus & 917\%  & 454\%  & 16\%  & 55\% & 78\% & 22.17 & 14.13 \\
\bestBenefit & 849\% & 426\% & 17\% & 54\% & 94\% & 22.17 & 14.62 \\
\highestDemand & 741\% & 371\% & 9\%  & 44\% & \textbf{89\%} & 23.44 & 8.43 \\
\chargerGreedy & 276\% & 185\% & 85\% & 81\% & 141\% & 22.75 & 34.0 \\
PCRL & \textbf{1016\%} & \textbf{497\%} & \textbf{6\%} & \textbf{26\%} & 92\%  & \textbf{21.40} & \textbf{0.83} \\
\midrule
\multicolumn{8}{c}{Dresden II}  \\
\cmidrule(lr){1-8}
\existingChargers & 100\% & 100\% & 100\% & 100\% & 100\% & 7.75 & 42.29 \\
\boundOpt & - & - & - & - & - & - & - \\
\boundOptPlus & 572\% & 296\% & 25\% & 79\% & 66\% & 7.75 & 24.41 \\
\bestBenefit & 542\% & 283\% & 25\% & 80\% & 71\% & 7.75 & 24.41 \\
\highestDemand & 420\% & 221\% & 15\% & 72\% & 62\% & 6.06 & 15.35 \\
\chargerGreedy & 226\% & 158\% & 92\% & 96\% & 109\% & 7.75 & 42.29\\
PCRL & \textbf{914\%} & \textbf{442\%} &  \textbf{3\%} & \textbf{28\%} & \textbf{56\%} & \textbf{3.51} & \textbf{0.35} \\
\bottomrule
\end{tabular}
}
\end{table*}

\subsection{Improving Real-World Charging Infrastructure}
\label{sec:eval:real}
In this experiment, we extend the existing real-world charging infrastructure by adding CS to the road network according to our PCRL approach. 
We compare our approach to all baselines described in Section \ref{sec:baselines}.
All considered models exhaust the available budget.
We evaluate all models according to the metrics in Section \ref{sec:setup:metrics}. 

\begin{figure*}
     \centering
     \begin{subfigure}[]{0.32\textwidth}
         \includegraphics[width=5.8cm]{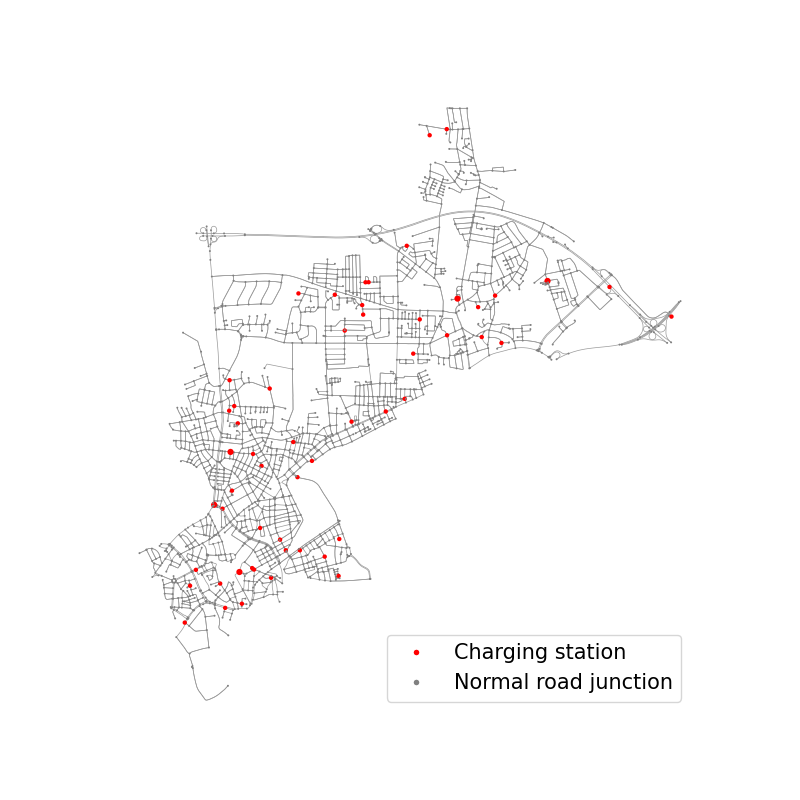}
         \caption{\existingChargers}
         \label{Fig:existingplan_Hannover}
     \end{subfigure}
     \hfill
     \begin{subfigure}[]{0.32\textwidth}
         \includegraphics[width=5.8cm]{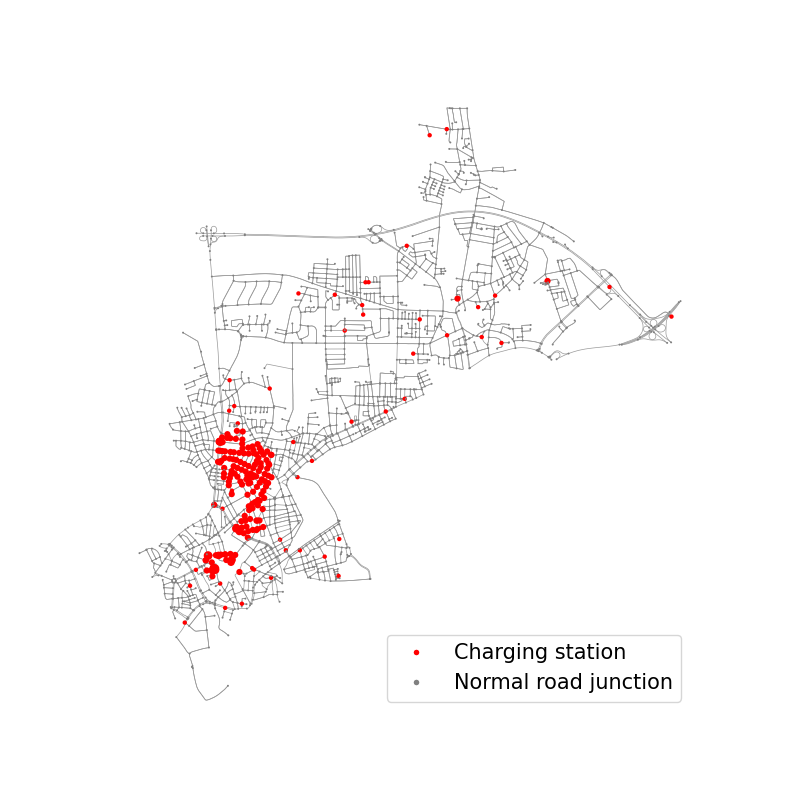}
         \caption{\boundOptPlus.}
         \label{Fig:baselineplus_Hannover}
     \end{subfigure}
     \hfill
     \begin{subfigure}[]{0.32\textwidth}
         \includegraphics[width=5.8cm]{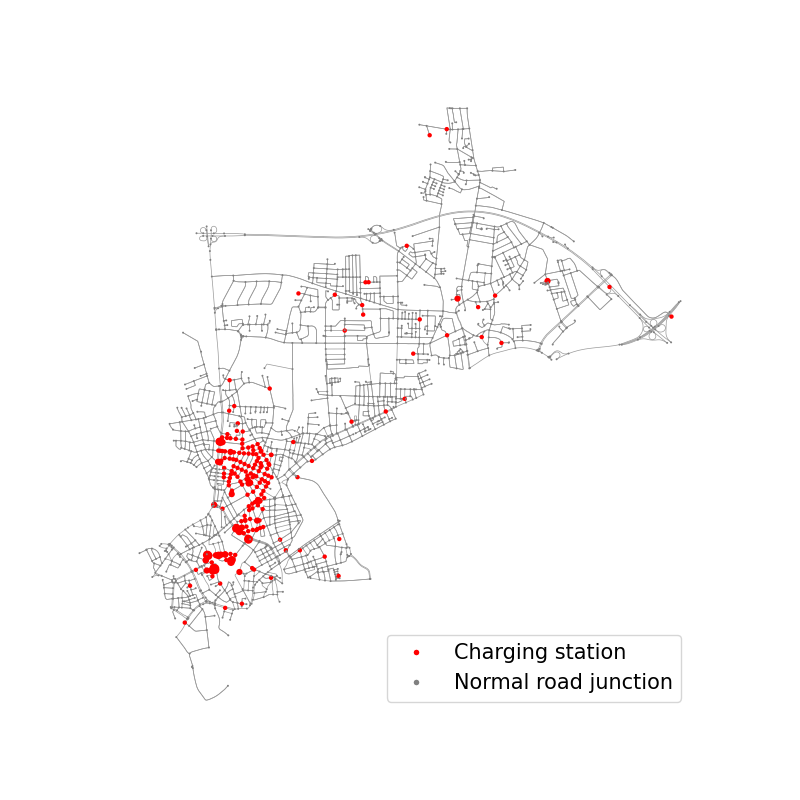}
         \caption{\bestBenefit.}
         \label{Fig:Benefitheuristic_Hannover}
     \end{subfigure}

     \begin{subfigure}[]{0.32\textwidth}
         \includegraphics[width=5.8cm]{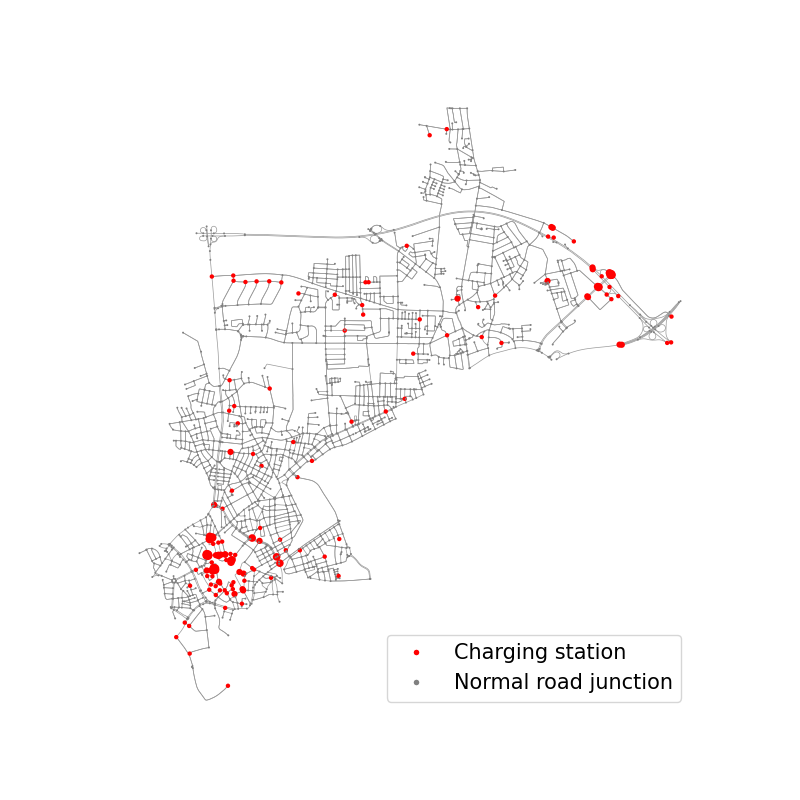}
         \caption{\highestDemand.}
         \label{Fig:Demandheuristic_Hannover}
     \end{subfigure}
     \hfill
     \begin{subfigure}[]{0.32\textwidth}
         \includegraphics[width=5.8cm]{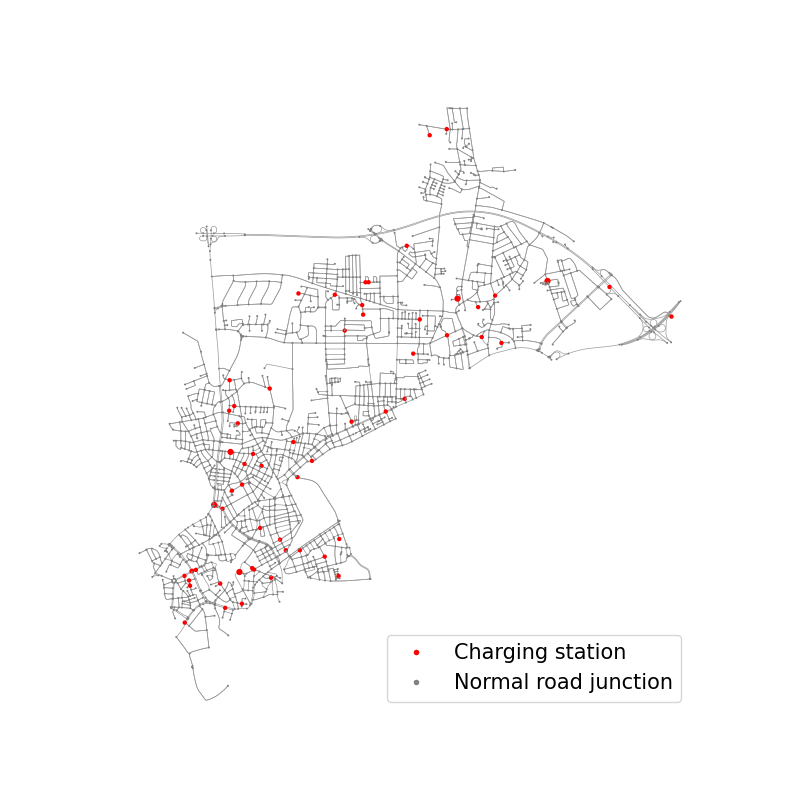}
         \caption{\chargerGreedy.}
         \label{Fig:KDD_Hannover}
     \end{subfigure}
     \hfill
     \begin{subfigure}[]{0.32\textwidth}
         \includegraphics[width=5.8cm]{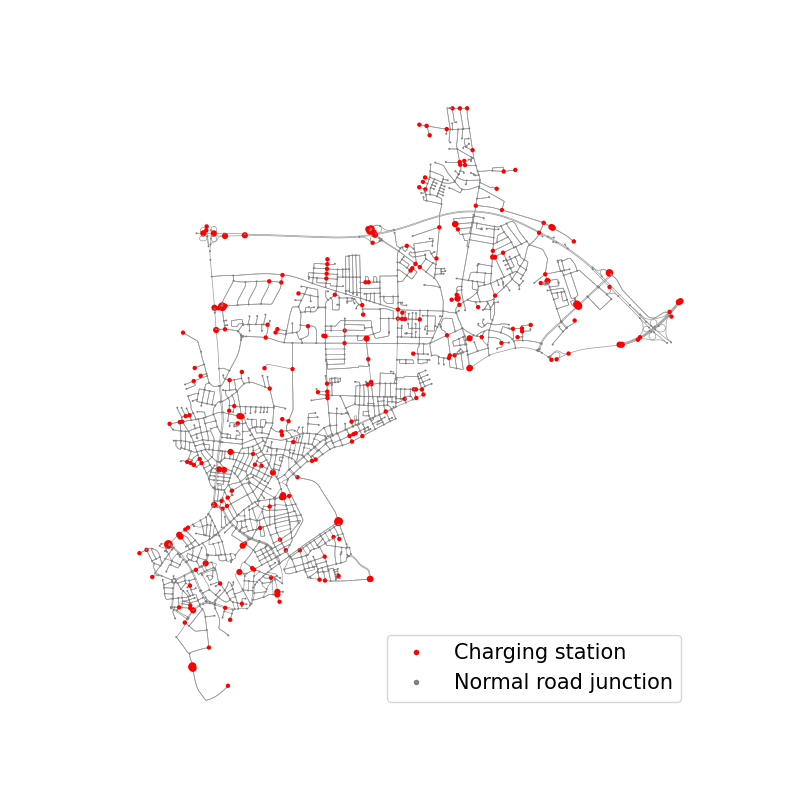}
         \caption{PCRL (ours).}
         \label{Fig:PCRL_Hannover}
     \end{subfigure}
    \caption{Charging plans after applying different algorithms on Hannover I. The thickness of the circle symbolises the capacity at the charging station.}
    \label{fig:HannoverI}
\end{figure*}

Table \ref{Tab:Hannover} presents the results for the datasets Hannover I and Hannover II, while Table \ref{Tab:Dresden} gives the results for Dresden I and Dresden II.
For Hannover I, we depict the resulting CP in Figure \ref{fig:HannoverI}. 

Our proposed PCRL model achieves the best results for all metrics on almost every dataset. 
In particular, PCRL achieves the best overall $score$ on every dataset.

%
% individual metrics
The remaining metrics indicate the individual aspects contributing to the $score$.
The $benefit$ quantifies the positive impact of the charging plan on the road network.
PCRL achieves the highest benefit on all datasets and therefore addresses the charging needs best.
Furthermore, PCRL outperforms the baselines regarding the travel time $travel$ in all datasets. 
Intuitively, a high benefit, i.e., a good supply with charging infrastructure, leads to lower travelling times for electric vehicle owners. 
Moreover, PCRL's waiting time $wait$, as well as the maximal waiting time $wait_{max}$, are the lowest for every dataset. 
In Dresden II (Table \ref{Tab:Dresden}), we can decrease the waiting time by 97\% indicating the high potential benefit of the effective distribution of CS over the road network.
Considering the charging time, PCLR achieves the best performance on Hannover I, Hannover II, and Dresden II and the second-best performance on Dresden I.
We observe that the \highestDemand baseline mainly uses the expensive charger types, resulting in the lowest charging time on Dresden I.
However, this strategy fails to address the charging demand of the entire road network resulting in the low overall $score$ of 741\% on Dresden I.
In contrast, our PCRL approach achieves a $score$ of 1016\% while losing only three percentage points charging time compared to \highestDemand.

Comparing the performance across datasets, we observe that the $score$s of the Dresden datasets are higher than for the Hannover datasets. 
Furthermore, we notice that the charging demand most strongly influences the density of the deployed stations.
We observe a more even distribution of Dresden's charging demand and estate price, allowing for better optimisation of charging and travel times, ultimately resulting in better plans.

% baselines
Considering the baselines, we observe that the \textsc{Bounding}\&\textsc{Op-} \textsc{timising}\xspace algorithm does not find a valid solution to the optimisation problem. 
Therefore, the \boundOpt is not applicable in real-world scenarios considered in this paper.
Our improved version of the baseline, \boundOptPlus, achieves the best score among the baselines but does not outperform our PCRL approach.
Figure \ref{fig:HannoverI} provides a visualisation of the charging plans computed by each model for the Hannover I dataset.
Each red circle represents a charging station; a higher radius indicates a higher capacity.
The baselines having the strategy to choose the CS with the potential highest benefit cluster the CS in the areas with the highest road junction density, as observed in Figure \ref{Fig:baselineplus_Hannover} and \ref{Fig:Benefitheuristic_Hannover}.
While these baselines choose positions with many neighbouring junctions, our PCRL approach ensures that each node in the network is provided with a nearby CS. Hence, the PCRL does not cluster the CS but distributes them over the road network (Figure \ref{Fig:PCRL_Hannover}). 
The more even distribution of charging stations is also reflected by the low waiting times achieved by PCLR.
Ultimately, the more balanced distribution of charging stations prevents the concentration of electric vehicles at a few overcrowded stations.

Figures \ref{Fig:Demandheuristic_Hannover} and \ref{Fig:KDD_Hannover} visualise the CS of the greedy baselines \highestDemand and \chargerGreedy that place the CS in locations with the highest demand or the potential benefit.
These baselines spend much budget on a few locations with the highest demand but neglect wide parts of the road network, resulting in poor benefit and travel time. 
In contrast, the PCRL illustrated in Figure \ref{Fig:PCRL_Hannover} generates smaller clusters distributed all over the road network. 
These CS with cheaper chargers satisfy the demand.
We observe a higher increase in benefit by placing many smaller charging stations than placing only one charging station with high capacity.

In summary, we observe that our PCRL model generates charging plans that ensure a broad supply with charging infrastructure while reducing the drivers' discomfort, e.g., long travel or waiting times.
Furthermore, the baselines fail to address the real-world needs, as they cluster CS or concentrate the capacity at a few locations only.

\subsection{Non-incremental Placement}
\label{sec:eval:new}
This experiment computes the charging plan from scratch without considering the existing charging infrastructure.
Table \ref{Tab:zerocase} shows the mean of the results for the four datasets.
Similar to the first experiment, the PCRL approach achieves the overall best score. 
We observe that all models besides \chargerGreedy{} achieve a higher score than the \existingChargers indicating that automated approaches can substantially improve the currently used strategies for planning charging infrastructure.
The \textsc{Bounding}\&\textsc{Op-}\textsc{timising}\textsc{+}\xspace baseline achieves the lowest charging time and waiting time.
However, this baseline primarily places expensive charger types on only a few charging stations. It neglects comprehensive parts of the road network, resulting in an even higher travel time than the \existingChargers.
The sparsity of charging stations renders this baseline unpractical for real-world scenarios.
In contrast, our \boundOptPlus approach maintains competitive charging and travel times while achieving the best travel time.

Comparing the non-incremental placement experiment with the experiment initiated with the real-world charging infrastructure, we observe an improvement of the mean score across all datasets from 683\% in the first experiment to 710\% in the second experiment.
The improvement indicates that PCLR can also improve the current charging infrastructure's positioning.

\begin{table*}[t]
\centering
\caption{Mean results for the non-incremental case. Evaluation metrics where higher values are better are marked with $\uparrow$. Metrics where lower values are better are labelled with $\downarrow$.  Best scores are marked bold.}
\label{Tab:zerocase}
\resizebox{310pt}{!}{%
\begin{tabular}{l c ccccccc} 
 \toprule
 \multicolumn{1}{c}{Algorithm} & \multicolumn{2}{c}{Score $\uparrow$} & \multicolumn{5}{c}{Cost $\downarrow$} \\
 \cmidrule(lr){2-3}\cmidrule(lr){4-8}
 & $score$ & $benefit$ & $wait$ & $travel$ & $charging$ &  $travel_{max}$ [min] & $wait_{max}$ [min]\\
\midrule
\existingChargers & 100\% & 100\% & 100\% & 100\% & 100\% & 14.56 & 32.79 \\
\boundOptPlus & 458\% & 247\% & \textbf{3\%} & 201\% & \textbf{30\%}  & 17.57 & 0.65 \\
\bestBenefit & 421\% & 231\% & 8\% & 182\% & 43\% & 17.50 & 2.14 \\
\highestDemand & 322\% & 178\% & 6\% & 127\% & 46\% & 14.50 & 1.50 \\
\chargerGreedy & 26\%  & 66\% & 78\% & 574\% & 64\% & 26.13 & 21.63 \\
PCRL  & \textbf{710\%} & \textbf{350\%} & 6\% & \textbf{38\%} & 66\%  & \textbf{11.24} & \textbf{0.59}\\
\bottomrule
\end{tabular}
}
\end{table*}

\subsection{Deployment of the PCRL}
\label{sec:eval:deployment}
In this section, we discuss the prerequisites for the deployment of the PCRL. The required datasets, see Section \ref{sec:Datasets}, comprise road network data and data regarding the existing charging infrastructure. These datasets are typically publicly accessible and easy to process.
Moreover, the PCLR requires data on the local estate prices and the density of detached houses often provided by city municipalities. 
This data is typically available to authorities planning traffic infrastructure.
Quantifying the electric vehicle charging demand is most challenging.
However, a wide range of datasets such as taxi trip data, floating car data, and routing data can serve as proxy information.
In general, we assume that charging demands correlate with travel demands.
Our open-source implementation of PCLR 
uses openly available libraries, i.e., Stable Baselines3 \cite{stablebaselines3}, OSMnx \cite{Boeing2017} and Gym \cite{gym}. 
We trained our approach on a CPU with eight cores within 24 hours for the four datasets used in this paper.
Our Q-learning approach has an observation space of size $|V| \times m$ and an action space of size 5.
Our approach is capable of building onto the existing charging infrastructure. PCLR can easily be applied in various scenarios, e.g., extending or relocating existing infrastructure, or planning charging infrastructure from scratch.

\section{Related Work}
\label{sec:relatedwork}

This section discusses related work in electric vehicle charging station placement and related optimisation algorithms.

In recent studies on the deployment of charging stations, the focus has been on different optimisation objectives: Liu et al. \cite{liu_optimal_2016} adopted a bilevel optimisation model to minimise the drivers' discomfort and maximise the gathered revenue. Vazifeh et al. \cite{vazifeh_optimizing_2019} aimed to cover the entire demand region while minimising drivers' travel times and the total number of charging stations by adopting a genetic algorithm. Liu et al. \cite{liu2019} set up a particle-swarm intelligent optimisation to minimise the CO\_2 emissions. Mourad et al. \cite{mourad_optimal_2020} formulated the optimisation problem, including alternative energy sources like photovoltaic and used a solver based on the simplex algorithm. Moreover, Sun et al. \cite{9632309} aimed to promote the usage of electric vehicles over conventional cars. Bae et al. \cite{9288870} included particular brand preferences into the optimisation.
In these studies, the authors optimised the location of the charging stations. However, the number of chargers per station remained constant, such that their approaches do not apply to our setting.

Another type of study considered optimising of the number of chargers per station. Du et al. \cite{10.1145/3219819.3220032} showed that the charger planning problem is NP-hard and proved a theoretical bound for their approximation algorithm. 
Cui et al. \cite{8673613} considered the number of chargers while including more practical aspects of urban systems, like the underlying power grid and its capacity. 

Gan et al. \cite{8950037} formulated a fast-charging station deployment problem in which different charger types were included. Meng et al. \cite{MENG2020120794} optimised the position and capacity of charging stations for electric taxis by adopting sequential construction planning. Another approach of Choi \cite{choi2020} formulated a large-scale charging station concept and optimised the charging infrastructure using a K-mean algorithm.
Although the previous studies considered an entire charging infrastructure solution, these studies did not integrate the existing charging stations. 
Zeng et al. \cite{ZengVW}, as well as Krallmann et al. \cite{8397185} extended the existing public infrastructure by using a genetic algorithm. Liu et al. \cite{10.1145/3347146.3359382} also examined the incremental case by using a greedy algorithm. 

We adopt the algorithms proposed in \cite{10.1145/3219819.3220032}, and \cite{10.1145/3347146.3359382} as baselines, as they represent recent approaches for charging station placement that consider road network topology as the most important aspect for charging station placement and are capable of extending the existing charging infrastructure.
Our approach optimises both positions and the number of chargers at the charging stations and extends the existing charging infrastructure. Furthermore, we include a novel element into the model of the charging placement problem, neglected so far to the best of our knowledge: The possibility of home charging electric vehicles that eases the problem of insufficient public charging infrastructure in urban areas \cite{ZHAO2020120039}.  

As the problem of charging station placement is NP-hard, approximation algorithms are adopted.  
Authors of previous studies employed different greedy algorithms \cite{10.1145/3219819.3220032}, \cite{10.1145/3347146.3359382}, \cite{9632309}, genetic algorithms \cite{ZengVW}, \cite{vazifeh_optimizing_2019}, and Bayesian Optimisation \cite{9288870} that either tend to cluster the nodes or neglect the cost-effectiveness. We can avoid this by combining different greedy strategies to a more refined policy by adopting reinforcement learning.  
Reinforcement learning has evolved as an essential method to solve sequential decision problems in a dynamic environment, and was recently adopted for recommending publicly accessible charging stations \cite{10.1145/3442381.3449934}, \cite{Tuchnitz}. 
In reinforcement learning, the agent seeks the best policy to solve a specific problem in the environment. The agent improves its strategy by getting rewards and punishments for the actions. 
 We adopt a deep Q network (DQN) reinforcement learning \cite{mnih2013playing} to make an agent learn an optimal strategy of finding the position of the charging station and their optimal charger type configuration.
To the best of our knowledge, our approach is the first attempt to adopt reinforcement learning for the optimal CS placement. 

\section{Conclusion}
\label{sec:conclusion}

In this paper, we investigated the problem of automatically determining the positions of electric vehicle charging stations in road networks.
To this end, we provide a novel formulation of the charging station placement problem that considers critical real-world requirements for determining efficient charging station positions, such as the possibility of charging electric vehicles at home.
Furthermore, we provide a novel reinforcement learning formulation of the charging station placement problem.
We conduct experiments on real-world road networks and determine possible extensions of the existing charging infrastructure. 
Our proposed approach, PCRL, achieves a benefit of up to 497\% compared to the existing charging infrastructure.
Moreover, we can reduce the maximum travel time to reach a charging station from approx. 8 minutes to 4 minutes.
The deployment of our approach requires typically openly available datasets such as road network data, charging station locations and census data. It also requires trip or routing data to estimate the charging demand.
Our findings provide evidence for the high potential use of our approach in deployed real-world scenarios. 
In future work, we plan to evaluate our charging plans with respect to the underlying power grid which is currently not included in our model.
Moreover, we would like to further refine our reinforcement learning model, e.g., by including deletions of charging stations into the action space or considering delayed rewards.

\begin{acks}
This work is partially funded by the Federal Ministry for Economic Affairs and Climate Action (BMWK), Germany under the projects ``d-E-mand'' (grant ID 01ME19009B) and ``CampaNeo'' (grant ID 01MD19007B), the European Commission (EU H2020, ``smashHit'', grant-ID 871477), and by the DFG, German Research Foundation (``WorldKG'', grant ID 424985896).
\end{acks}

\balance

\bibliographystyle{ACM-Reference-Format}
\bibliography{ref}

\end{document}